\RequirePackage{fix-cm}
\documentclass[twocolumn]{svjour3}          
\smartqed  

\usepackage[utf8]{inputenc} 
\usepackage[T1]{fontenc}    
\usepackage{url}            
\usepackage{booktabs}       
\usepackage{amsfonts}       
\usepackage{nicefrac}       
\usepackage{microtype}      
\usepackage{xcolor}         
\usepackage{soul}
\usepackage{color, xcolor}
\soulregister{\cite}7
\usepackage{epsfig}
\usepackage{graphicx}
\usepackage{amsmath}
\usepackage{amssymb}
\usepackage{multirow}
\usepackage{color, colortbl}
\usepackage{subcaption}
\usepackage{subfloat}
\usepackage{tabularx}
\usepackage[export]{adjustbox}
\usepackage{threeparttable}
\usepackage{pifont}
\usepackage[misc]{ifsym} 
\usepackage{algorithm}
\usepackage{algpseudocode}
\usepackage{paralist}
\usepackage{listings} 
\usepackage{amssymb}
\usepackage{makecell}
\usepackage{bbding}
\usepackage{caption}
\usepackage{url}
\makeatletter
\newcommand\footnoteref[1]{\protected@xdef\@thefnmark{\ref{#1}}\@footnotemark}
\makeatother

\definecolor{COLOR_CSID}{HTML}{e0f5ff}
\definecolor{COLOR_NEAROOD}{HTML}{ffefe0}
\definecolor{COLOR_FAROOD}{HTML}{ffdebf}
\definecolor{COLOR_MEAN}{HTML}{f0f0f0}

\usepackage{xspace}
\usepackage{comment}
\usepackage{bm}
\usepackage{wrapfig}

\usepackage[colorlinks,linkcolor=red,anchorcolor=blue,citecolor=blue,CJKbookmarks=True]{hyperref}

\definecolor{citecolor}{HTML}{0071BC}
\definecolor{linkcolor}{HTML}{ED1C24}

\renewcommand\paragraph{
  \@startsection{paragraph} 
  {4} 
  {\z@} 
  {.5em \@plus1ex \@minus.2ex} 
  {-1.5em} 
  {\normalfont\normalsize\bfseries} 
}
\DeclareRobustCommand\onedot{\futurelet\@let@token\@onedot}
\def\@onedot{\ifx\@let@token.\else.\null\fi\xspace}

\def\eg{\emph{e.g.}} 
\def\ie{\emph{i.e.}}

\def\etal{\emph{et al.}}
\makeatother

\begin{document}

\captionsetup[figure]{labelfont={bf}}
\sloppy 


\title{Skim then Focus: Integrating Contextual and Fine-grained Views for Repetitive Action Counting}
\author{Zhengqi Zhao$^{1\dag}$ \and 
        Xiaohu Huang$^{12\dag}$ \and
        Hao Zhou$^{2{~\textrm{\Letter}}}$ \and
        Kun Yao$^{2}$ \and
        Errui Ding$^{2}$ \and
        Jingdong Wang$^{2}$ \and
        Xinggang Wang$^{1}$ \and
        Wenyu Liu$^{1}$ \and
        Bin Feng$^{1{~\textrm{\Letter}}}$
}
        
\institute{
    1. Huazhong University of Science and Technology, Wuhan, China. \\
    2. Department of Computer Vision Technology (VIS), Baidu Inc.  \\
    $\dag$ Contributed equally to this work. \\
    ${~\textrm{\Letter}}$ Corresponding authors: fengbin@hust.edu.cn and zhouh156@mail.ustc.edu.cn.\\
    }
    
\date{Received: date / Accepted: date}


\maketitle

\begin{abstract}
The key to action counting is accurately locating each video's repetitive actions.
Instead of estimating the probability of each frame belonging to an action directly, we propose a dual-branch network, i.e., SkimFocusNet, working in a two-step manner. 
The model draws inspiration from empirical observations indicating that humans typically engage in coarse skimming of entire sequences to grasp the general action pattern initially, followed by a finer, frame-by-frame focus to determine if it aligns with the target action.
Specifically, SkimFocusNet incorporates a \textit{skim branch} and a \textit{focus branch}. 
The \textit{skim branch} scans the global contextual information throughout the sequence to identify potential target action for guidance. 
Subsequently, the \textit{focus branch} utilizes the guidance to diligently identify repetitive actions using a long-short adaptive guidance (LSAG) block.
Additionally, we have observed that videos in existing datasets often feature only one type of repetitive action, which inadequately represents real-world scenarios.
To more accurately describe real-life situations, we establish the Multi-RepCount dataset, which includes videos containing multiple repetitive motions.
On Multi-RepCount, our SkimFoucsNet can perform \textit{specified action counting}, that is, to enable counting a particular action type by referencing an exemplary video.
This capability substantially exhibits the robustness of our method.
Extensive experiments demonstrate that SkimFocusNet achieves state-of-the-art performances with significant improvements. We also conduct a thorough ablation study to evaluate the network components. The source code will be published upon acceptance.

\end{abstract}

\section{Introduction}
\label{sec:intro}
Performing periodic movements over time leads to the formation of repetitive actions, a phenomenon prevalent in daily activities such as physical training. 
In computer vision, counting repetitive actions in videos is a newly arisen topic, which has potential applications in intelligent systems, \eg, intelligent workout tracking \cite{kong2022human}. 
The key step of action counting is to detect the action periods along the time dimension accurately, therefore learning to differentiate action-related and -unrelated frames is critical for this task.

\begin{figure}[t]
    \centering
    \includegraphics[width=\linewidth]{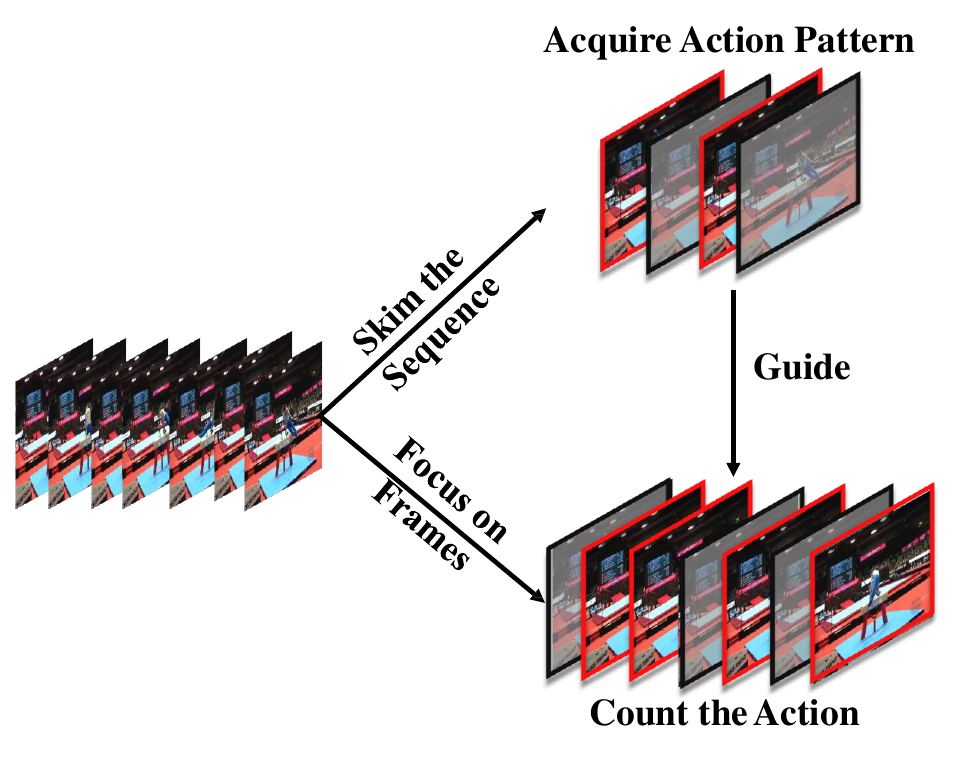}
    \caption{Illustration of the life experience that humans count actions in two steps: (1) We skim the sequence coarsely and acquire the possible target action pattern as guidance. (2) With the guidance of the skim process, we focus on the frames that include the target action to conduct counting.}
    \label{fig:motivation}
\end{figure}

The previous top-performing methods tackle this task in different ways.
Zhang \etal \cite{zhang2020context} establish a context-aware and scale-insensitive refinement method to locate the periods. 
Differently, RepNet \cite{dwibedi2020counting} and TransRAC \cite{hu2022transrac} construct correlation matrices \cite{junejo2010view,vaswani2017attention,vlachos2005periodicity} to mine the relations among frames.
Furthermore, TransRAC \cite{hu2022transrac} estimates the period distribution in the form of density maps, which are widely adopted in crowd counting \cite{liu2019context,ranjan2018iterative,tan2019crowd,wan2019adaptive,zhang2016single,zhang20223d,zhang2022wide,xu2022autoscale}.
We notice that these methods rely heavily on identifying the similarity between frames but lack comprehension of the target action. 
However, if we do not know which action is the target action that needs to be counted, the feature learned could be ambiguous especially when various interference actions exist in the video.
For example, when doing physical training, one may combine multiple action classes to build a workout plan.
In summary, we argue that there is an important idea that has long been neglected: \textit{we need to determine which action to count before starting counting.} 

This argument is consistent with human life experience. 
As shown in Fig.\ref{fig:motivation}, humans first skim the whole sequence to quickly locate the frames of the possible target action to acquire an action pattern, since there are a lot of irrelevant frames in the video as distractions. Then, taking the located frames as guidance, humans can focus on the frames with large similarities and dedicatedly count the actions. Naturally, this human perception procedure can be summarized as a \textit{skim-then-focus} process. In this way, the \textit{skim} process provides cues about the target action to the \textit{focus} stage, which helps filter out the irrelevant frames and exploit the informative ones, eventually benefiting action counting. Compared with the \textit{skim} process exploring the global context, the \textit{focus} stage keeps an eye on local video segments to notice the fine-grained visual clues.

Inspired by the above observation, we propose a network for repetitive action counting, dubbed \textit{SkimFocusNet}. The core idea of SkimFocusNet is to mimic the human perception mechanism for locating fine-grained clues under the guidance of contextual information across the sequence. Specifically, we construct a network with a dual-branch architecture, where each branch corresponds to the \textit{skim} and \textit{focus} processes, respectively. The \textit{skim branch} takes a long contextual sequence as input for quick viewing and samples a short instructive sequence from it. As for the \textit{focus branch}, it takes the short instructive sequence and the local fragments of the video as inputs. The instructive sequence provides information on the target action while local fragments deliver fine-grained details with a high temporal resolution. They are both beneficial for conducting meticulous counting. Afterward, the \textit{focus branch} generates an informative feature vector using the instructive sequence which represents the properties of the target action in the whole sequence. This vector is then used as a guide, based on which the feature expression of each frame from the local fragment is adjusted dynamically, \ie, highlights the action-relevant frames, and suppresses action-irrelevant frames. The adjustment is completed through a long-short adaptive guidance (LSAG) block, which further fuses the features along the temporal dimension with long-term and short-term time scales for enriching temporal representation diversities.

Moreover, the videos within current datasets solely feature one single type of repetitive action, failing to accommodate scenarios where multiple repetitive actions coexist.
To better reflect complicated situations, we construct the Multi-RepCount dataset with the data unit of a video containing various repetitive actions and an exemplary video for reference.
Based on that, we introduce a novel problem setting, namely \textit{specified action counting}, which refers to counting occurrences of a specified action within a video, utilizing an exemplary video as a reference.
Conducting \textit{specified action counting} on the Multi-RepCount dataset provides an effective means of evaluating the robustness of various methods under complex scenarios.

We evaluate SkimFocusNet on three datasets, \ie, RepCount \cite{hu2022transrac}, UCFRep \cite{zhang2020context}, and the proposed Multi-RepCount, where RepCount and UCFRep typically include only one type of repetitive action for each video. SkimFocusNet sets new state-of-the-art performance on the three datasets and outperforms the previous methods by a large margin. 

Summarily, the contribution can be concluded as follows:
\begin{itemize}
\item We propose a dual-branch framework (SkimFocusNet) for action counting, where the \textit{skim branch} provides informative cues for the \textit{focus branch}, and helps it pay delicate attention to action-related frames.

\item We propose a long-short adaptive guidance (LSAG) module for adaptively conducting the guiding from the \textit{skim branch} to the \textit{focus branch} in detail. LSAG further explores multi-scale modeling for enhancing temporal representation.

\item We propose a new problem setting of \textit{specified action counting} and construct a new dataset Multi-RepCount, which is effective for comparing the robustness of different methods. 
 
\item Extensive experiments conducted on three datasets, RepCount \cite{hu2022transrac}, UCFRep \cite{zhang2020context} and the proposed Multi-RepCount, demonstrate the superior performance of SkimFocusNet. Additionally, further ablation experiments are conducted to prove the effectiveness of the proposed components in the network.
\end{itemize}


\section{Related Works}
\subsection{Action Counting}
Our work focuses on action counting in videos that involve repetitive actions. 
Previous classical approaches transform the video into a one-dimensional time-domain signal \cite{albu2008generic,azy2008segmentation,cutler2000robust,laptev2005periodic,lu2004repetitive,panagiotakis2018unsupervised,pogalin2008visual,tralie2018quasi,tsai1994cyclic}. 
Early researchers then estimate the period of the original video sequence by utilizing the Fourier transform \cite{azy2008segmentation,briassouli2007extraction,cutler2000robust,pogalin2008visual}, which is an effective mathematical tool for processing periodic information from a one-dimensional signal.
However, this mathematical method assumes that the motion is stable and continuous \cite{burghouts2006quasi,chetverikov2006motion,davis2000categorical,thangali2005periodic}, which may not hold in real-life scenarios \cite{runia2019repetition,zhang2020context}.

Levy and Wolf \cite{levy2015live} introduce an online classification network trained on a synthetic dataset to estimate action periods.
The fixed length of the input sequence for their network may hinder performance when dealing with original sequences of varying lengths.
Ruina \etal \cite{runia2018real} propose a method adopting wavelet transform to handle the non-stationary video sequence. 
This method is incapable of well generalizing to real-world datasets due to complexity.
Zhang \etal \cite{zhang2020context} propose an iterative refinement framework for action counting, which is not friendly to computational cost.
Dwibedi \etal \cite{dwibedi2020counting} construct a self-similarity matrix among frames of the input sequence to achieve class-agnostic action counting. 
However, it predefines the range of actions in each video, limiting its ability to handle sequences with a higher number of actions.
Zhang \etal \cite{zhang2021repetitive} incorporate multi-modal information in action counting, which integrates visual signals with sound for the first time.
Hu \etal \cite{hu2022transrac} propose multi-scale input and use the transformer blocks to predict the density maps of action periods.
Li \etal \cite{Li_2024_WACV} propose a two-branch framework incorporating RGB and motion to enhance the foreground motion feature learning.

In this paper, we point out a key concept that we need to skim the sequence in advance to locate the possible target action. However, the previous methods all tend to estimate the action periods directly but are not meant to take an early look at the sequence to determine which action to count in general. This may lead to inaccurate attention to those irrelevant frames. 
In contrast, we propose a dual-branch architecture, where a \textit{skim branch} first captures the noteworthy frames as guidance, then a \textit{focus branch} counts the action based on the guidance in detail. 

\subsection{Action Recognition}
In a closely related field, \ie, action recognition, several dual-branch architectures are very well known, \eg, Two-Stream networks \cite{simonyan2014two} and SlowFast \cite{feichtenhofer2019slowfast}. Two-stream networks \cite{simonyan2014two} take RGB images and optical flows as inputs for each branch, and finally, ensemble the two-stream predictions for recognition. Differently, SlowFast \cite{feichtenhofer2019slowfast} feed sequences with different frame rates into two different paths, aiming at capturing spatial semantics and temporal motion simultaneously. 

Our method differs from them in 1) Unlike Two-Stream networks \cite{simonyan2014two}, we only use images as inputs instead of multi-modal information. 2) Unlike Two-Stream networks \cite{simonyan2014two} and SlowFast \etal \cite{feichtenhofer2019slowfast}, the two branches in our framework are given nonequivalent status. Exactly, the \textit{skim branch} is more like an assistant to the \textit{focus branch}.
\begin{figure*}[t]
  \centering
  \includegraphics[width=\linewidth]{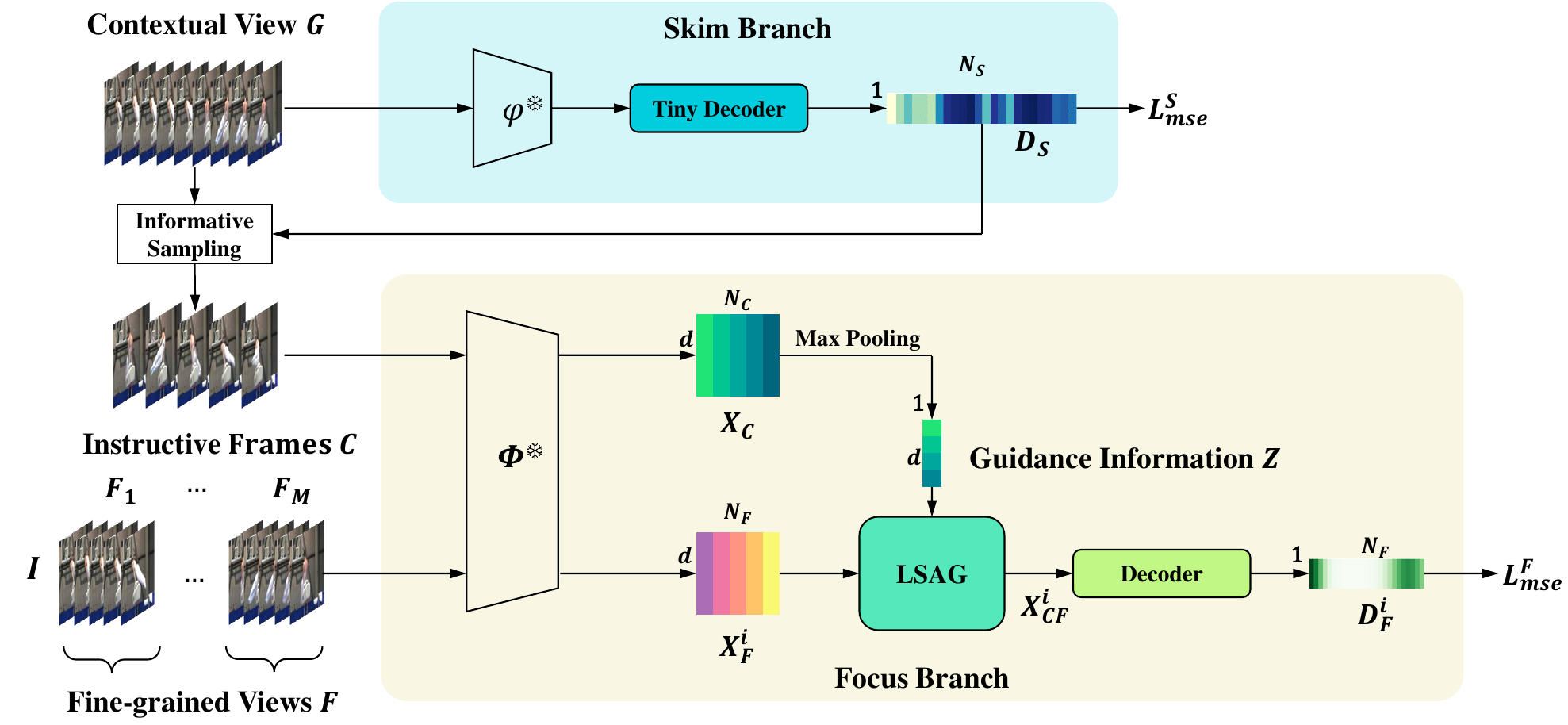}
   \caption{Framework overview. There are two branches in SkimFocusNet, \ie, the \textit{skim branch} and the \textit{focus branch}. The contextual view $G$ is processed by the \textit{skim branch} which is a lengthy sequence aimed at capturing as much of the entire video as possible. Next, an informative sampling module is employed to sample the instructive frames $C$ for the \textit{focus branch}. The instructive frames $C$ and the fine-grained views $F$ are passed through the \textit{focus branch} and encoded as feature $X_C$ and $X_F^i$ respectively. Max pooling is applied to extract critical guidance information $Z$ from feature $X_C$. The long-short adaptive guidance (LSAG) block integrates it with the feature $X_F^i$ to help differentiate the action-relevant and -irrelevant features. Mean Square Error (MSE) loss is utilized to supervise the learning process of the two branches separately.}
   \label{fig: framework}
\end{figure*}

\section{SkimFocusNet} 
In this section, we first describe the architecture of SkimFocusNet, and then elaborate on the proposed components in detail.

\subsection{Architecture}
\label{sec:arch}
Fig.\ref{fig: framework} illustrates the overall framework of SkimFocusNet.
SkimFocusNet incorporates a \textit{skim branch} and a \textit{focus branch}, both consisting of an encoder and a decoder that adhere to the overall framework established by TransRAC \cite{hu2022transrac}.
The encoder $\mit\Phi$ in the \textit{focus branch} adopts a complete video swin transformer, whereas the encoder $\varphi$ in the \textit{skim branch} utilizes only the first layer of the video swin transformer.
The decoder in the \textit{focus branch} employs a correlation matrix using a self-attention block and then applies a transformer-based period predictor to output a vector.
Meanwhile, the tiny decoder in the \textit{skim branch} follows the same design but with fewer computational layers.
As the objective of the \textit{skim branch} is to capture the action pattern coarsely, it possesses a comparatively lighter network structure than the \textit{focus branch}.

Specifically, given a video $I$ including multiple repetitive actions, after downsampling by a factor $R$, we truncate it into a contextual view $G$ with the length $N_S$ and divide it into $M$ fine-grained views $F=\{F_{i}|i=1,2,\cdots, M\}$ with the same length $N_F$.

The \textit{skim branch} takes the contextual view $G$ as an input. 
Through encoder $\varphi$, and a tiny decoder, the \textit{skim branch} outputs the confidence map $D_S=[\alpha^1,\alpha^2,\cdots,\alpha^{N_S}]$, where $\alpha^n$ denotes the predicted confidence value that the $n$-th frame belongs to the target action.
According to the confidence map $D_S$, the \textit{skim branch} samples the instructive frames $C$ of length $N_C$ from the contextual view $G$ through the informative sampling module, which is discussed in \ref{sec:index}.
The instructive frames $C$ contain fewer frames but more representative information about the target action which is beneficial for counting.

The \textit{focus branch} takes the instructive frames $C$ and one fine-grained view $F_i$ from $F$ as inputs. 
They are fed into an encoder $\mit\Phi$ to produce the feature embedding $X_C \in \mathbb{R} ^ {N_C \times d}$ and $X_F^i \in \mathbb{R} ^ {N_F \times d}$, respectively, where $d$ denotes the feature dimension of each frame.
We apply the max pooling layer on the feature embedding $X_C$ to acquire guidance information $Z \in \mathbb{R} ^ {d}$.
Next, the \textit{focus branch} utilizes the LSAG block to integrate the feature embedding $X_F^i$ with the guidance information $Z$. 
The LSAG block first adaptively adjusts feature embedding according to $Z$ and then it models long-term and short-term relations in the time domain to adapt to actions with different cycle lengths. The specific structure of LSAG is detailed in Sec.\ref{sec: LSAG}.
Afterward, the \textit{focus branch} outputs the density map $D_F^i=[\beta^1,\beta^2,\cdots,\beta^{N_F}]$ which describes the distribution of action periods, where $\beta^n$ denotes the predicted density value for the $n$-th frame.
Particularly, the sum of $D_F^i$ along the time dimension denotes the predicted number of repetitions for the fine-grained view $F_i$.
During inference, we can get the predicted number of repetitions for the whole video by summing up the outputs from all of the $M$ fine-grained views.
\begin{equation}
    L=L_{\text{mse}}^{S}+L_{\text{mse}}^{F}
    \label{eq: loss}
\end{equation}

Finally, we use Mean Square Error (MSE) as the loss function to penalize the distance between the predicted and ground-truth density maps for both the \textit{skim branch} and the \textit{focus branch}.
As shown in Eq.\ref{eq: loss}, the overall loss function $L$ is the sum of the \textit{skim branch} loss $L_{mse}^S$ and the \textit{focus branch} loss $L_{mse}^F$.

\subsection{Informative Sampling} \label{sec:index}
\begin{figure}[t]
  \centering
  \includegraphics[width=0.63\linewidth]{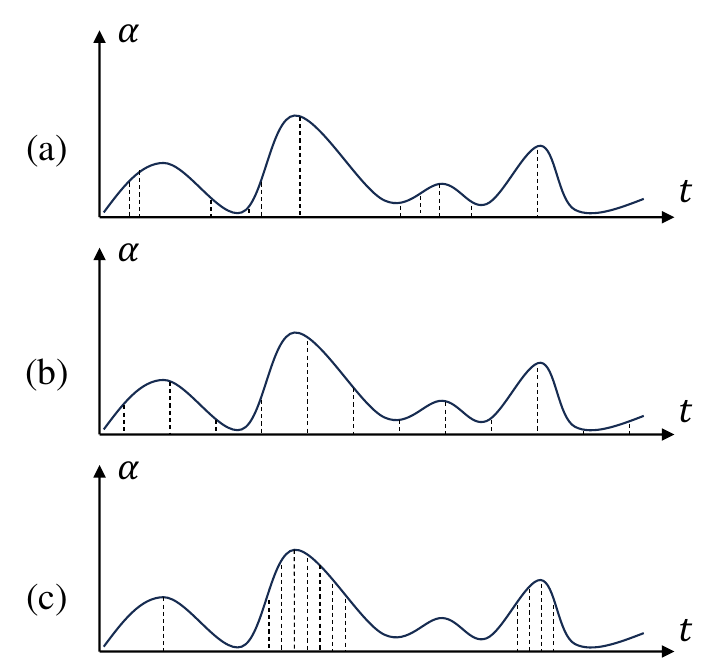}
  \caption{The implementations of different informative sampling strategies. It illustrates the sampling results of the (a) random sampling strategy, (b) uniform sampling strategy, and (c) top $N_C$ sampling strategy.}
  \label{fig: index}
\end{figure}

The informative sampling module is designed to extract the instructive frames $C$ from the \textit{skim branch}.
As shown in Fig.\ref{fig: index}, this section shows a couple of implementations of the informative sampling strategies, \ie random sampling, uniform sampling, and top $N_C$ sampling.

The random sampling strategy doesn't rely on the predicted confidence map $D_S$ and the sampling frames are likely to cover the whole input sequence.
It serves as the baseline strategy for comparison.
The uniform sampling strategy evenly distributes the sampling frames in the temporal domain, capturing the overall information extracted from the contextual view $G$.
The top $N_C$ sampling strategy selects $N_C$ frames with the highest $N_C$ values in the $D_S$ map, which contain the most discriminative information of the contextual view $G$.

According to the performance comparison of different informative sampling strategies in Sec.\ref{sec: ablation}, we choose the top $N_C$ sampling strategy for SkimFocusNet.

\subsection{LSAG} \label{sec: LSAG}
\begin{figure}[t]
  \centering
  \includegraphics[width=0.63\linewidth]{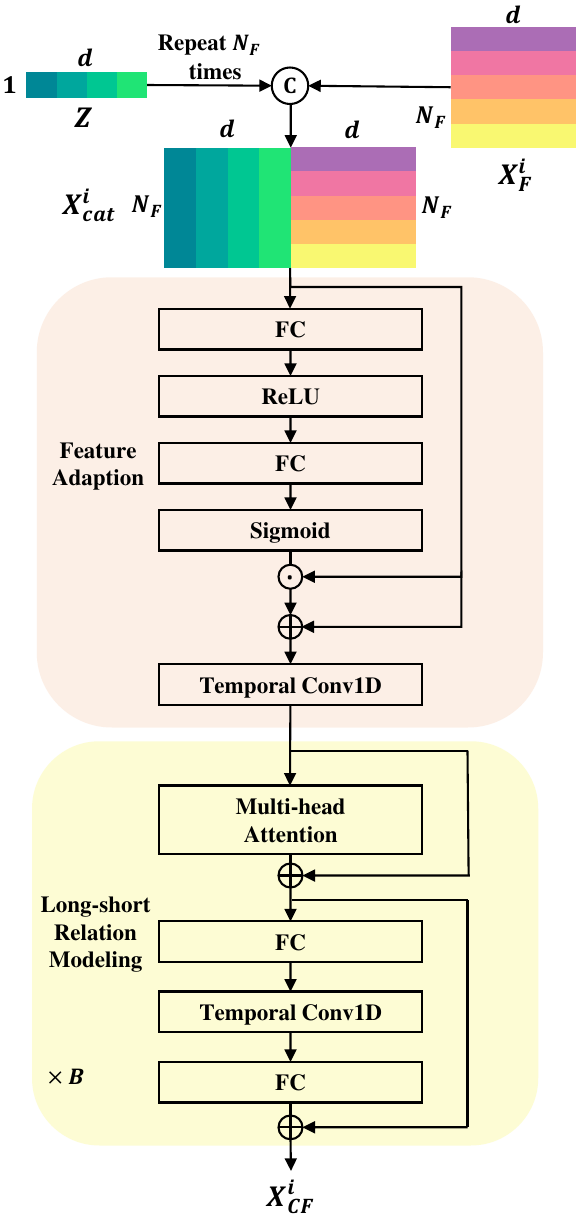}
  \caption{The long-short adaptive guidance (LSAG) block integrates critical guidance information $Z$ with fine-grained feature embedding $X_F^i$.}
  \label{fig: LSAG}
\end{figure}

Fig.\ref{fig: LSAG} shows the overall architecture of the LSAG block. It takes the guidance information $Z$ with the \textit{focus branch} feature embedding $X_F^i$ as inputs and then outputs the enriched feature embedding $X_{CF}^i$. $X_{CF}^i$ retains fine-grained temporal resolution and contains instructive information from the \textit{skim branch}. 

Specifically, the LSAG block consists of two components: feature adaption and long-short relation modeling. The feature adaption is to emphasize the frames containing repetitive actions and ignore the irrelevant ones and backgrounds, according to the guidance from the \textit{skim branch}. Long-short relation modeling models temporal features in different scales, which can better understand the motion pattern of different actions with both short periods and long periods.

For feature adaption, we first expand the guidance information $Z$ in the time domain and then concatenate it with the feature embedding $X_F^i$, which produces the combined embedding $X_{cat}^i \in \mathbb{R}^{N_F \times 2d}$. The process can be formulated as Eq.\ref{eq: cat}:
\begin{equation}
    X_{cat}^i=cat(X_F^i, repeat(Z,N_F)),
    \label{eq: cat}
\end{equation}
where $repeat(Z,N_F)$ denotes expanding the guidance information $Z$ in the time domain for $N_F$ times.

Then, the feature embedding $X_{cat}^i$ goes through a bottleneck with a sigmoid activation function to extract a set of attention in the channel dimension. Next, we use 1D temporal convolutions to aggregate the feature embedding of the attention output with a shortcut connection.
In this way, we can adaptively adjust feature representation to emphasize the frames that correspond to repetitive actions and ignore those containing background movements.

Next, we use $B$ long-short relation modeling blocks to model long- and short-term relations along the sequence.
The multi-head self-attention layer is used to model long-term relations like contextual information.
After that, a 1D convolution layer is applied to capture short-term relations in adjacent frames.
Finally, the LSAG block produces the feature embedding $X_{CF}^i$, which contains temporal features in different scales and thus can better capture the motion pattern of different actions with both long and short periods. In this way, $X_{CF}^i$ is a reliable feature embedding to predict the repetition number for the fine-grained view $F_i$.
\begin{figure}[t]
  \centering
  \includegraphics[width=0.9\linewidth]{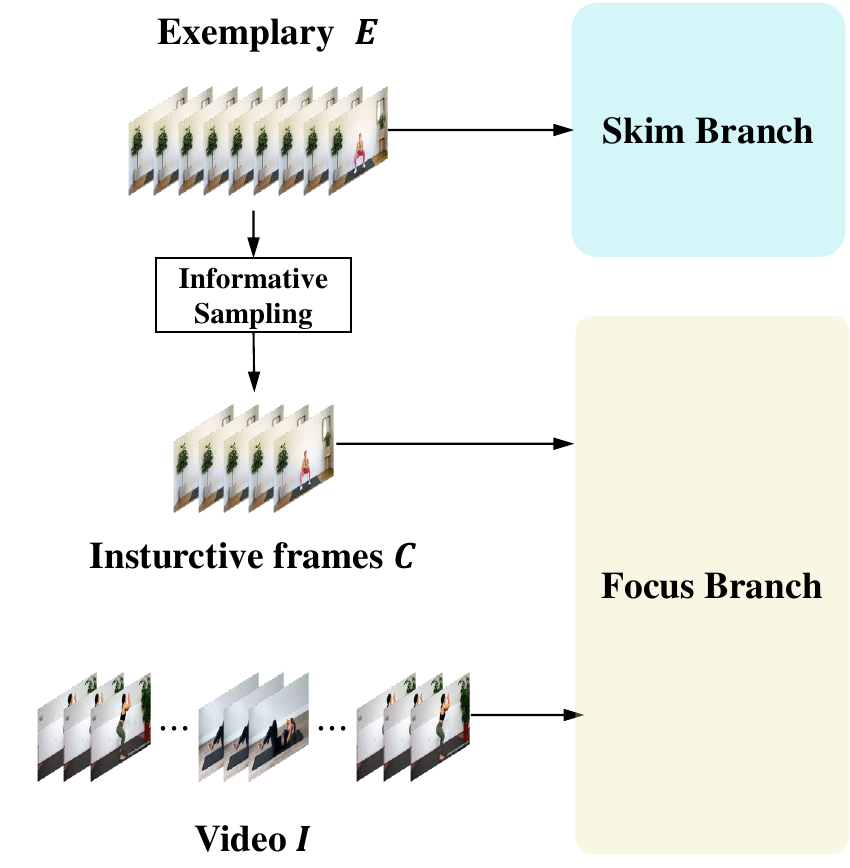}
  \caption{The example of performing \textit{specified action counting} using SkimFocusNet.}
  \label{fig: specified}
\end{figure}

\section{Specified Action Counting}
Repetitive actions performed by humans tend to exhibit great complexity in real-life scenarios.
For instance, engaging in diverse exercises such as push-ups or sit-ups is a common practice during workouts.
However, current datasets for repetitive action counting, such as RepCount \cite{hu2022transrac}, primarily focus on a single type of repetitive action, with minimal interruptions such as breaks between repetitions.
These interruption actions are characterized by their brevity and lack of repetition and barely disrupt the network’s similarity-based counting process.
To better reflect real-life scenarios, we introduce a new problem setting of specified action counting and construct a more intricate dataset Multi-RepCount.

\subsection{Problem Setting}
We propose a new problem setting of \textit{specified action counting} to simulate counting problems under complicated scenarios.
In detail, given a video $I$ and an exemplary video $E$, our objective is to quantify the occurrence of the target action $a$ in video $I$ which comprises multiple repetitive actions by comparing it with the repetitive action $a$ depicted in exemplary video $E$.
Even though the other actions may repeat themselves in the video $I$, they are regarded as background movements concerning the target action $a$.
In conclusion, the main goal of the proposed \textit{specified action counting} is to accurately count the number of the specified actions in the videos containing different categories of repetitive actions.
However, without specifying the target action, it is very challenging for class-agnostic action counting to perform well under such circumstances.
Therefore, besides the video with different actions, another video of a certain repetitive action is provided as an exemplar.

Notably, the proposed \textit{SkimFocusNet} can readily handle this new setting.
As shown in Fig.\ref{fig: specified}, the \textit{Skim branch} takes the exemplary video $E$ as input and outputs the instructive frames $C$.
Then, the \textit{focus branch} takes the instructive frames $C$ and a fine-grained fragment of video $I$ as inputs to perform \textit{specified action counting}.
With the help of the guidance information, we believe that our method can differentiate whether the action is of the target category or not.

\subsection{Multi-RepCount}
Based on the dataset RepCount \cite{hu2022transrac}, we create a more complex dataset Multi-RepCount, where each video contains multiple types of repetitive actions.

Statistically, the RepCount dataset includes 9 categories of repetitive actions including \textit{others}.
We discard the \textit{others} category because it does not refer to a specific action category and, therefore, is not applicable in this setting.
When composing Multi-RepCount, we first select 3 videos separated from each category of the training set as candidate exemplary video set.
Then, for each video of a certain target action $a$, we insert clips from other categories as distractions for each video.
The insertion position and the order of the clips are randomly generated.
In this way, we compose the video $I$ containing multiple actions.
Additionally, the frames of the target action $a$ take up half of the percentage of the entire video $I$.
Moreover, from the candidate exemplary video set, we randomly choose one exemplary video $E$ for the corresponding category $a$ associated with video $I$, representing the fundamental data unit. 

\begin{figure}[t]
    \centering \includegraphics[width=0.45\textwidth]{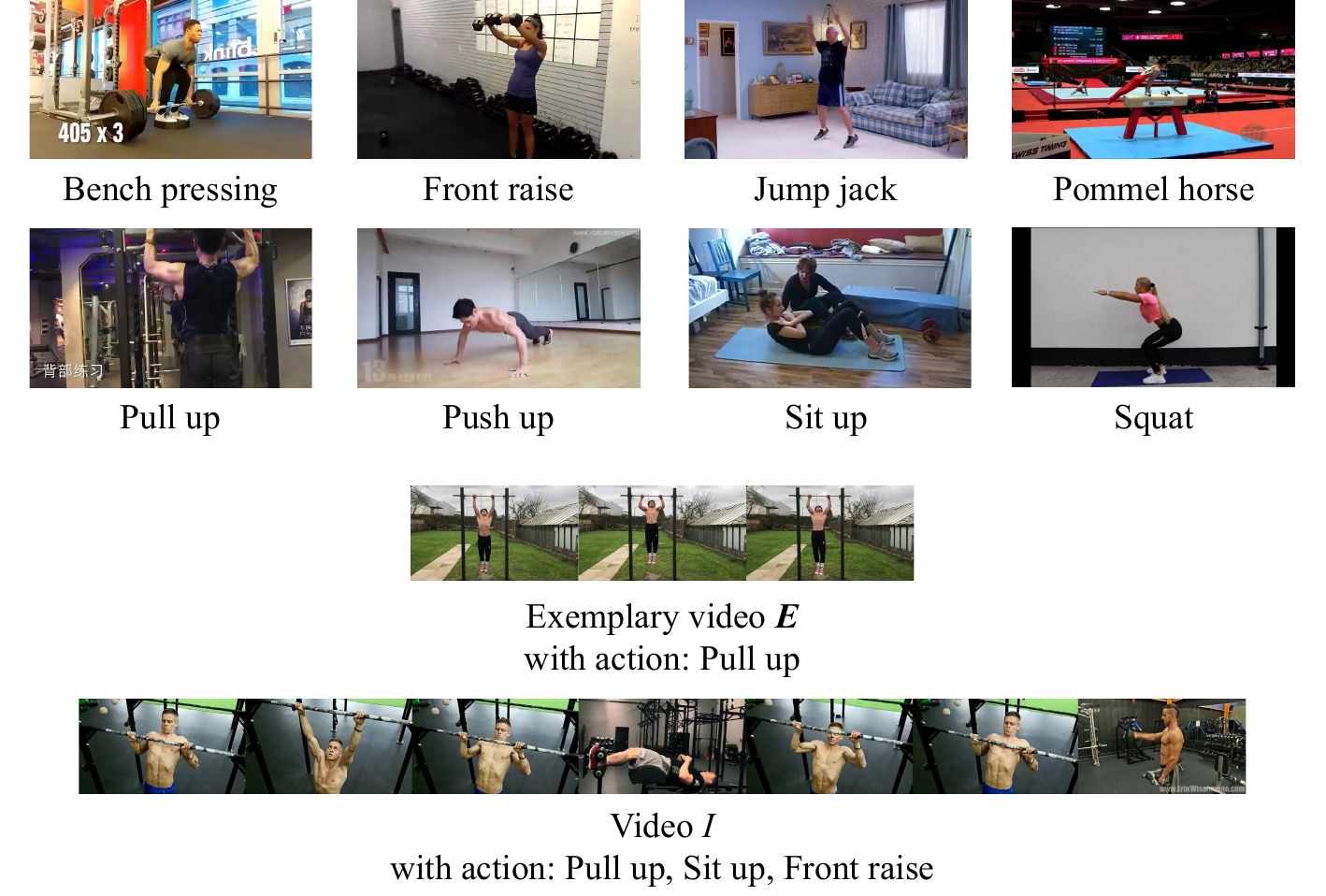}
    \caption{The 8 classes of the repetitive actions existed in the Multi-RepCount and an example data unit of Multi-RepCount.}
    \label{fig: dataset}
\end{figure}

In conclusion, our Multi-RepCount has 2 main modifications compared to RepCount.
(1) each video contains repetitive actions from multiple categories.
(2) each video is packed with a randomly selected exemplary video.
Figure \ref{fig: dataset} shows the example data unit of the dataset Multi-RepCount.

\section{Experiments}
\subsection{Evaluation Metrics}

\noindent \textbf{Off-By-One (OBO) Accuracy}. OBO will be accumulated when the absolute difference between the prediction and the ground truth is no larger than one. 
Otherwise, OBO will not be accumulated, which is formulated as:

\begin{equation}
    \text{OBO}=\frac{1}{N}\sum_{i=1}^{N}[|\hat{c_i} - {c_i}|\leq{1}],
    \label{eq: OBO}
\end{equation}

where $\hat{c_i}$, $c_i$, and $N$ denote the prediction, the ground truth, and the number of the test sequences, respectively. In this way, a higher OBO indicates better performance. \\

\noindent \textbf{Mean Absolute Error (MAE)}. MAE measures the absolute error between the prediction and the ground truth. Therefore, a lower MAE indicates better performance. The MAE calculation can be denoted as:

\begin{equation}
    \text{MAE}=\frac{1}{N} \sum_{i=1}^{N}\frac{|\hat{c_i}-{c_i}|}{c_i}.
    \label{eq: MAE}
\end{equation}
\subsection{Datasets}
There are many existing datasets for repetition counting \cite{hu2022transrac,zhang2020context,dwibedi2020counting,levy2015live,runia2018real}.
However, due to their limited size, some early proposed datasets \cite{levy2015live,runia2018real} are not suitable for training complex networks.
As a result, we evaluate our network on the two large-scale widely used datasets and the proposed Multi-RepCount.

\noindent{\textbf{RepCount Part-A.}} The RepCount Part-A dataset \cite{hu2022transrac} contains 1041 videos with fine-grained annotations and significant variation in length. 
These videos are collected from the YouTube website.
While containing many anomaly cases, it has a larger average count number and a longer average duration than any other dataset.
Due to the above characteristics, it becomes the most challenging one.

\noindent{\textbf{UCFRep.}} The UCFRep dataset \cite{zhang2020context} contains 526 videos of 23 categories.
It comprises the annotated repetition videos from the action recognition dataset UCF101 \cite{soomro2012ucf101}.

\noindent{\textbf{Multi-RepCount.}} The proposed Multi-RepCount dataset is a synthetic dataset made up of repetitive action clips of RepCount.
The basic data unit of the dataset is a video pair consisting of a video for counting and an exemplary video for reference. 
Statistically, Multi-RepCount contains 984 videos of 8 repetitive action classes and each video contains more than 3 kinds of different actions.
\subsection{Implementation Details}
We implement our method with the PyTorch platform and train it with an NVIDIA GeForce RTX3090 GPU.
We train the network (except for the pre-trained VideoSwin Transformer or ResNet) for an overall 200 epochs and set the batch size as 8. Unless otherwise stated, VideoSwin Transformer is used by default. Besides, we apply the Adam optimizer with a declining learning rate of $8 \times 10^{-6}$. 

The frame lengths for the contextual view $G$ are set to 256.
The frame lengths for the instructive Frames $C$ and the \textit{focus branch} $N_F$ are set to 32 and 64.
The downsampling rate ($R$) for both inputs for the \textit{focus branch} and the contextual view $C$ is set as 4. The number of long-short relation modeling blocks ($B$) is set as 3.

\subsection{Comparison with State-of-the-art Methods}

\noindent \textbf{RepCount Part-A.} Tab.\ref{tab:RepCount} shows the performance comparison between SkimFocusNet and previous competitive methods,
including action counting methods (RepNet, Zhang \etal, and TransRAC, Li \etal), action recognition methods (X3D, TANet, and Video SwinT), and action segmentation methods (Huang \etal and ASFormer). We find that SkimFocusNet outperforms these methods by a large margin. 
Compared with the best competitor, our SkimFocusNet decreases MAE from $0.3841$ to $0.2489$ and improves OBO from $0.3860$ to $0.5166$, which are relatively improved by $35.2\%$ and $33.8\%$, respectively.
The comparison of RepCount Part-A strongly suggests the superior counting accuracy of the proposed method.

\noindent \textbf{UCFRep.} Following the evaluation protocol in TransRAC \cite{hu2022transrac}, we first train the models on the RepCount Part-A dataset, then test them on the UCFRep dataset. In Tab.~\ref{tab: UCF}, we compare SkimFocusNet with RepNet, Zhang \etal, TransRAC, and Li \etal, from which we can notice that SkimFocusNet still significantly outperforms all of them.
However, the improvement is not as substantial as the RepCount Part-A which indicates the difference between the two datasets and the difficulty of such an evaluation protocol.
The comparison on UCFRep proves the generality of the proposed method.

\noindent \textbf{Multi-RepCount.} 
Table \ref{tab: joint} shows the performance comparison between previous competitive methods.
However, both TransRAC and RepNet are single-stream networks that cannot utilize critical information from the exemplary video $E$.
Therefore, besides our SkimFocusNet, we test our model without the \textit{skim branch} on the Multi-RepCount for a fair comparison.
In general, it is reasonable for single-stream methods to count the main repetitions since the target action $a$ takes up half of the percentage.
Without the guidance information from the other branch, we find that SkimFocusNet still outperforms other methods in the single-stream framework.
With the critical information provided by the \textit{skim branch}, the performance of SkimFocusNet is improved by a large margin which suggests the importance of the skim process under complex situations.
The comparison on Multi-RepCount proves the robustness of the proposed method.
\begin{table}[t]
    \centering
    \caption{
    Performance of different methods on RepCount part-A test set when trained on the train set of RepCount. * denotes
that the results are based on our implementation.
    }
    \small
    \begin{tabular}{l|cccc}
    \toprule
    Method & \multicolumn{1}{c}{MAE $\downarrow$} & OBO $\uparrow$   \\ \midrule
    X3D \cite{feichtenhofer2020x3d}  & \multicolumn{1}{c}{0.9105}  & 0.1059 \\
    TANet \cite{liu2021tam}& \multicolumn{1}{c}{0.6624}          & 0.0993           \\
    Video SwinT \cite{liu2022video}    & \multicolumn{1}{c}{0.5756}          & 0.1324           \\ 
    Huang \etal \cite{huang2020improving}    & \multicolumn{1}{c}{0.5267}          & 0.1589           \\ 
    ASFormer* \cite{yi2021asformer} & \multicolumn{1}{c}{0.5279}          & 0.2980\\ \midrule
    RepNet \cite{dwibedi2020counting}         & \multicolumn{1}{c}{0.9950}          & 0.0134           \\
    Zhang \etal \cite{zhang2020context}    & \multicolumn{1}{c}{0.8786}          & 0.1554           \\
    TransRAC \cite{hu2022transrac}       & \multicolumn{1}{c}{0.4431}          & 0.2913            \\
    Li \etal \cite{Li_2024_WACV}       & \multicolumn{1}{c}{\underline{0.3841}}          & \underline{0.3860}            \\
    \textbf{SkimFocusNet (ours)}        & \multicolumn{1}{c}{\textbf{0.2489}} & \textbf{0.5166} \\ \bottomrule
    \end{tabular}
    \label{tab:RepCount}
\end{table}

\begin{table}[t]
    \centering
    \caption{Performance of different methods on UCFRep when trained on RepCount part-A.}
    \small
    \begin{tabular}{l|cc} \toprule
    Method & MAE $\downarrow$  & OBO $\uparrow$ \\ \midrule
    RepNet \cite{dwibedi2020counting} & 0.9985 & 0.0090    \\
    Zhang \etal \cite{zhang2020context}  & 0.5913 & 0.3046 \\
    TransRAC \cite{hu2022transrac}  & 0.6401  & 0.3240  \\
    Li \etal \cite{Li_2024_WACV}  & \underline{0.5227}  & \underline{0.3500}  \\
    \textbf{SkimFocusNet (ours)}  &\textbf{0.5023}  &\textbf{0.3905}  \\
    \bottomrule  
    \end{tabular}
    \label{tab: UCF}
\end{table}

\begin{table}[h]
      \centering
     \small
     \caption{Performance of different methods on Multi-RepCount.}
		\begin{tabular}{c|cc}
			\toprule
			Method                   & \multicolumn{1}{c}{MAE $\downarrow$} & OBO  $\uparrow$ \\ \midrule
			RepNet \cite{dwibedi2020counting}            & \multicolumn{1}{c}{0.7521}          & 0.1600          \\
                TransRAC \cite{hu2022transrac}          & \multicolumn{1}{c}{0.6860}          & 0.1800         \\
                SkimFocusNet \textit{w/o} \textit{skim branch}          & \multicolumn{1}{c}{\underline{0.6014}}        & \underline{0.2133}          \\
			\textbf{SkimFocusNet (ours)}                    & \multicolumn{1}{c}{\textbf{0.3514}}          & \textbf{0.3400}         \\
            \bottomrule
		\end{tabular}
	\label{tab: joint}
\end{table}

\subsection{Complexity Analysis for Action Counting Methods}
In Tab. \ref{tab:complexity analysis and performance}, we analyze the complexity and performance of different methods on the RepCount dataset. 
We train all methods for 200 epochs and the test-time batch size is set to 1 for all methods. 
Besides, for RepNet \cite{dwibedi2020counting} and Zhang \etal \cite{zhang2020context}, we use the same frame-sampling strategy as TransRAC \cite{hu2022transrac}. 

From Tab. \ref{tab:complexity analysis and performance}, we find that: 
(1) With a bit more parameters and longer training time, our method can achieve the best performance compared with others. The extra parameters are due to the dual-branch design and the extra training time is due to the basic training data of video fragments.
(2) Compared to the previous method TransRAC \cite{hu2022transrac}, we improve the inference time from 38.02 seconds to 18.12 seconds which is due to the abandonment of the sliding window.
Moreover, during inference, each contextual view is processed only once for all the fine-grained views in the video which greatly improves efficiency.

\begin{table}[t]
    \centering
    \small
    \caption{Complexity analysis and performance on RepCount using an NVIDIA 3090 GPU. Total Parameters (ToParam, M). Training time (TT, hour). Inference time (IT, second). Notably, \textbf{the training time} is measured by using the same training settings (except for frame sampling) for all methods, and \textbf{the inference time} is measured on the whole test set.}
    \resizebox{\linewidth}{!}{
    \begin{tabular}{c|ccccc}
    \toprule
       Method & ToParam  &TT  &IT   &MAE$\downarrow$ &  OBO$\uparrow$ \\\midrule
        RepNet \cite{dwibedi2020counting} & 42.74 & 8.03 & 7.25 & 0.9950& 0.0134  \\
        Zhang \etal \cite{zhang2020context}&47.60 & 148.32 & 115.48 & 0.8786 &0.1554  \\
        TransRAC \cite{hu2022transrac}& 42.28 & 12.53 & 38.02 & 0.4431 &0.2913  \\
        \textbf{Ours} & 61.50 & 43.21 & 18.12 & \textbf{0.2489} &\textbf{0.5166}  \\\bottomrule
    \end{tabular}}
    
    \label{tab:complexity analysis and performance}
\end{table}

\subsection{Ablation Study} \label{sec: ablation}
In this section, we conduct extensive diagnostic experiments to evaluate the effectiveness of our model design. 

\begin{table}[b]
    \centering
    \caption{Study of the effectiveness of the proposed components on the RepCount Part-A dataset.}
    \small
    \setlength{\tabcolsep}{6pt}{
    \begin{tabular}{ccc|cc}
        \toprule
        \multicolumn{3}{c|}{Componet} & \multicolumn{2}{c}{RepCount A} \\
        \textit{Focus} & \textit{Skim} & LSAG & MAE $\downarrow$ & OBO $\uparrow$ \\\midrule
        $\checkmark$ & & & 0.3161 & 0.3709 \\
        $\checkmark$ & $\checkmark$ & &0.2794 &0.4238 \\
        $\checkmark$ & & $\checkmark$ &0.2901 & 0.3974\\
        $\checkmark$ & $\checkmark$ & $\checkmark$ & \textbf{0.2489} & \textbf{0.5166} \\
        \bottomrule
    \end{tabular}
    }
    \label{tab:component}
\end{table}

\noindent \textbf{Effectiveness of the proposed components.} In Tab.\ref{tab:component}, we evaluate the effectiveness of the proposed components in SkimFocusNet. 
We use basic CNN to replace LSAG in certain experiments.
From the results, it can be noticed that: (1) In the first experiment, without \textit{skim branch} and LSAG, the performance drops by a large margin. (2) By comparing the 2nd and the 3rd experiments, we find that, based on the \textit{focus branch}, the \textit{skim branch} contributes more improvement than LSAG, demonstrating the importance of the skimming process before counting. (3) Using all three components can achieve the best performance, revealing these components play complementary roles to each other.

\begin{table}[t]
    \centering
    \small
    \caption{Study of the proposed dual-branch design by replacing the encoder and decoder with different networks on the RepCount Part-A dataset.}
    \resizebox{\linewidth}{!}{
    \begin{tabular}{l|cc|cc}
    \toprule
    Framework & Encoder & Decoder & MAE $\downarrow$ & OBO $\uparrow$ \\\midrule
    RepNet \cite{dwibedi2020counting} & ResNet & Classification & 0.9950 & 0.0134 \\
    TransRAC \cite{hu2022transrac} & VideoSwin & Density-map & 0.4431 & 0.2913 \\\midrule
    \multirow{4}{*}{\textbf{\makecell[c]{Ours\\(Dual-branch)}}} & ResNet & Classification & 0.5968 & 0.1987 \\
    & ResNet & Density-map &0.5339 &0.2317 \\
    & VideoSwin & Classification & 0.2689 & 0.4834 \\
    & VideoSwin & Density-map & \textbf{0.2489} & \textbf{0.5166} \\\bottomrule
    \end{tabular}
    }
    \label{tab:encoder_decoder}
\end{table}

\noindent \textbf{Versatility of the dual-branch design.} 
The \textit{skim-then-focus} architecture is proposed to seamlessly integrate with various encoders and decoders. To verify this, in Tab.\ref{tab:encoder_decoder}, we implement the encoders and decoders with different networks. 
The first two dual-branch experiments based on encoder ResNet perform better than ResNet-based RepNet.
For the VideoSwin encoder, we can find the last two experiments show dual-branch method has better performance than the VideoSwin-based TransRAC.
The experiments strongly support our dual-branch design philosophy.

\noindent\textbf{Number of the downsampling rate $R$.}
As shown in Tab.\ref{tab:R}, we conduct experiments to compare the performance of different downsampling rates $R$ (2, 4, and 8).
From the results, we observe that a downsampling rate that is either too high or too low is not beneficial to performance. Therefore, we set $R=4$ for optimal performance.

\begin{table}[h]
     \centering
     \small
     \caption{Study of the impact of the downsampling rate $R$ in fine-grained view acquisition on the RepCount Part-A dataset, in terms of MAE and OBO metrics.}
		\begin{tabular}{c|cc}
			\toprule
			Method                   & \multicolumn{1}{c}{MAE $\downarrow$} & OBO  $\uparrow$ \\ \midrule
			$R$=2            & \multicolumn{1}{c}{0.2781}          & 0.4702        \\
			$R$=4           & \multicolumn{1}{c}{\textbf{0.2489}}          & \textbf{0.5166}          \\
   $R$=8                     & \multicolumn{1}{c}{0.2980}          & 0.4437 \\ 
            \bottomrule
		\end{tabular}
	\label{tab:R}
\end{table}

\noindent \textbf{Impact of the informative sampling strategies.}
In Tab.\ref{tab: index select}, we investigate the impact of the informative sampling strategies.
From the results, we can find that:
(1) Compared with the random sampling strategy, the results of the uniform sampling are similar which indicates the extracted general information brings almost no improvement.
(2) Our method can achieve the best performance when using top $N_C$ sampling which suggests significant information is superior to general information.

\begin{table}[h]
    \centering
    \caption{Study of informative sampling strategy on the RepCount Part-A dataset.}
    \small
    \setlength{\tabcolsep}{10pt}{
    \begin{tabular}{c|cc} \toprule
        Method  & MAE $\downarrow$ & OBO  $\uparrow$ \\ \midrule
	Random sampling     &  0.2570     & 0.4503  \\
    Uniform sampling     &  0.2526     & 0.4636          \\
    Top $N_C$ sampling     &  \textbf{0.2489}     & \textbf{0.5166}  \\\bottomrule
    \end{tabular}
    }
    \label{tab: index select}
\end{table}

\begin{table}[ht]
 \centering
 \caption{
		Study of the sampling frames length $N_C$ for instructive frames $C$ on the RepCount Part-A dataset, in terms of the MAE and OBO metrics.}
		\begin{tabular}{c|cc}
			\toprule
			Method                & \multicolumn{1}{c}{MAE $\downarrow$} & OBO  $\uparrow$ \\ \midrule
			$N_C=0$         & \multicolumn{1}{c}{0.3161}          & 0.3709          \\
			$N_C=16$   & \multicolumn{1}{c}{0.2537}          & 0.4768          \\
			$N_C=32$   & \multicolumn{1}{c}{0.2489}          & \textbf{0.5166}           \\
			$N_C=64$   & \multicolumn{1}{c}{\textbf{0.2421}}   & 0.5099          \\
			$N_C=128$   & \multicolumn{1}{c}{0.2670}   & 0.4702          \\
            $N_C=256$   & \multicolumn{1}{c}{0.2629}   & 0.4503          \\
            \bottomrule
		\end{tabular}
	\label{tab:skim ablation}
\end{table}

\noindent \textbf{Number of sampling frames for instructive frames $C$.} 
In Tab.\ref{tab:skim ablation}, we evaluate the effectiveness of the \textit{skim branch} on the RepCount Part-A dataset. 
From the results, several noteworthy conclusions are summarized: 
(1) Compared with not using the \textit{skim branch} ($N_C = 0$), using only 16 frames for the \textit{skim branch} can effectively improve the performance. It reveals that even if we skim the sequence with a limited number of frames, it can still provide useful guidance to locate the action periods. 
(2) When we sample more frames for the \textit{skim branch}, the performance improves obviously, which demonstrates that the abundant contextual clues can indeed contribute to action counting. 
(3) When we further feed more frames to the \textit{skim branch}, it is found that the performance declines. One possible explanation is that over-sampling brings increasing noise, which degrades the quality of guidance information. To achieve a better trade-off between computational cost and performance, we choose to feed 32 frames to the \textit{skim branch}.

\noindent \textbf{Number of the fine-grained view length $N_F$.}
To investigate the effects of the fine-grained view length $N_F$, we conducted experiments with $N_F$ values of 32, 64, and 128. As shown in Table \ref{tab:N}, setting the fine-grained view length $N_F$ to 64 outperforms other experiments in terms of the OBO metric. Additionally, the first two settings achieve similar performances in terms of the MAE metric. However, setting $N_F$ to 32 increases training time consumption, and setting it to 128 takes up more GPU memory. To achieve a better trade-off between computational cost and performance, we set the fine-grained view length $N_F$ to 64.
\begin{table}[h]
     \centering
     \small
     \caption{Study of the fine-grained view length $N_F$ on the RepCount Part-A dataset, in terms of MAE and OBO metrics.}
		\begin{tabular}{c|cc}
			\toprule
			Method                   & \multicolumn{1}{c}{MAE $\downarrow$} & OBO  $\uparrow$ \\ \midrule
			$N_F$=32           & \multicolumn{1}{c}{0.2565}          & 0.4702         \\
			$N_F$=64           & \multicolumn{1}{c}{\textbf{0.2489}}          & \textbf{0.5166}          \\
			$N_F$=128                     & \multicolumn{1}{c}{0.3025}          &   0.3974        \\
            \bottomrule
		\end{tabular}
	\label{tab:N}
\end{table}

\begin{table}[h]
    \centering
    \small
    \caption{Study of the effectiveness of the LSAG module design on the RepCount dataset.}
    \setlength{\tabcolsep}{2pt}{
    \begin{tabular}{l|cc}  \toprule
    Method & MAE $\downarrow$ & OBO  $\uparrow$ \\ \midrule
    LSAG \textit{w/o} Feature Adaption & 0.2559      & 0.4437  \\
    LSAG \textit{w/o} Long-short Relation Modeling   & 0.3111  & 0.4040   \\
    LSAG   & \textbf{0.2489} & \textbf{0.5166}  \\ \bottomrule
    \end{tabular}
    }
    \label{tab: LSAG ablation}
\end{table}

\noindent \textbf{Impact of the module design in LSAG.} 
In Tab.\ref{tab: LSAG ablation}, we investigate the impact of the module design in LSAG. 
From the results, we can notice that: 
(1) Without the feature adaption block, counting performance declines. 
It indicates that it is important to adapt the feature representation of each frame in the \textit{focus branch} according to the guidance from the \textit{skim branch}. 
(2) Removing the long-short relation modeling degrades the performance obviously, which shows that richer granularities of temporal expression would benefit the motion representation and help better capture the characteristics of different actions. 


\begin{figure}[t]
    \centering
    \includegraphics[width=\linewidth]{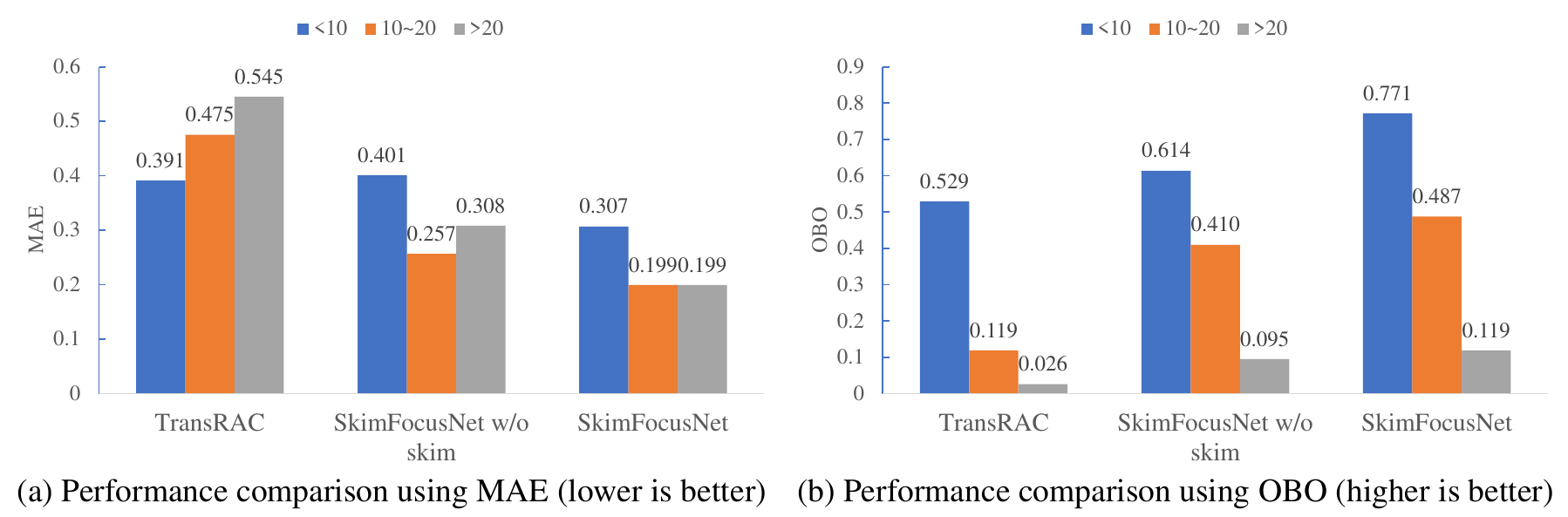}
    \caption{Performance comparison in different ranges of action counts on the RepCount Part-A dataset.}
    \label{fig:barplot}
\end{figure}

\begin{table}[h]
    \centering
    \caption{Study of the number ($B$) of the long-short relation modeling blocks in LSAG on the RepCount Part-A dataset.}
    \small
    \setlength{\tabcolsep}{10pt}{
    \begin{tabular}{c|cc} \toprule
        Method  & MAE $\downarrow$ & OBO  $\uparrow$ \\ \midrule
	B=1     &  0.2699     & 0.4172          \\
	B=3     &  \textbf{0.2489}     & \textbf{0.5166}  \\
	B=5     &  0.2828  & 0.4437    \\  \bottomrule
    \end{tabular}
    }
    \label{tab: number B}
\end{table}

\noindent \textbf{Number of the long-short relation modeling blocks in LSAG.} 
In Tab.\ref{tab: number B}, we investigate the impact of the value of $B$. By increasing $B$ from 1 to 3, the performance is improved significantly (especially on OBO), which is brought by richer temporal representation with more blocks. When we further increase $B$ to 5, the performance declines on both metrics. In this paper, we set $B$ as 3 for a trade-off between computational cost and performance.

\noindent \textbf{Performance comparison in different ranges of action counts.} In Fig.\ref{fig:barplot}, we present the performance comparison in different ranges of counts on the RepCount Part-A dataset. Several conclusions are summarized: 
(1) Our method achieves the best performance within all ranges, proving its superior capacity under different scenarios. 
(2) With the \textit{skim branch}, our method can consistently improve its performance in various situations. 
(3) The performances of all methods decline in terms of the OBO metric when the number of actions in a video increases, which is somehow reasonable since such scenarios are relatively harder. But when it comes to videos with more than 20 actions, the performances become terrible due to the accumulated errors. 
It reveals a potential direction for future research in this field, \ie, solving dense repetitive action counting in long videos.

\begin{figure*}[t]
  \centering
  \includegraphics[width=\linewidth]{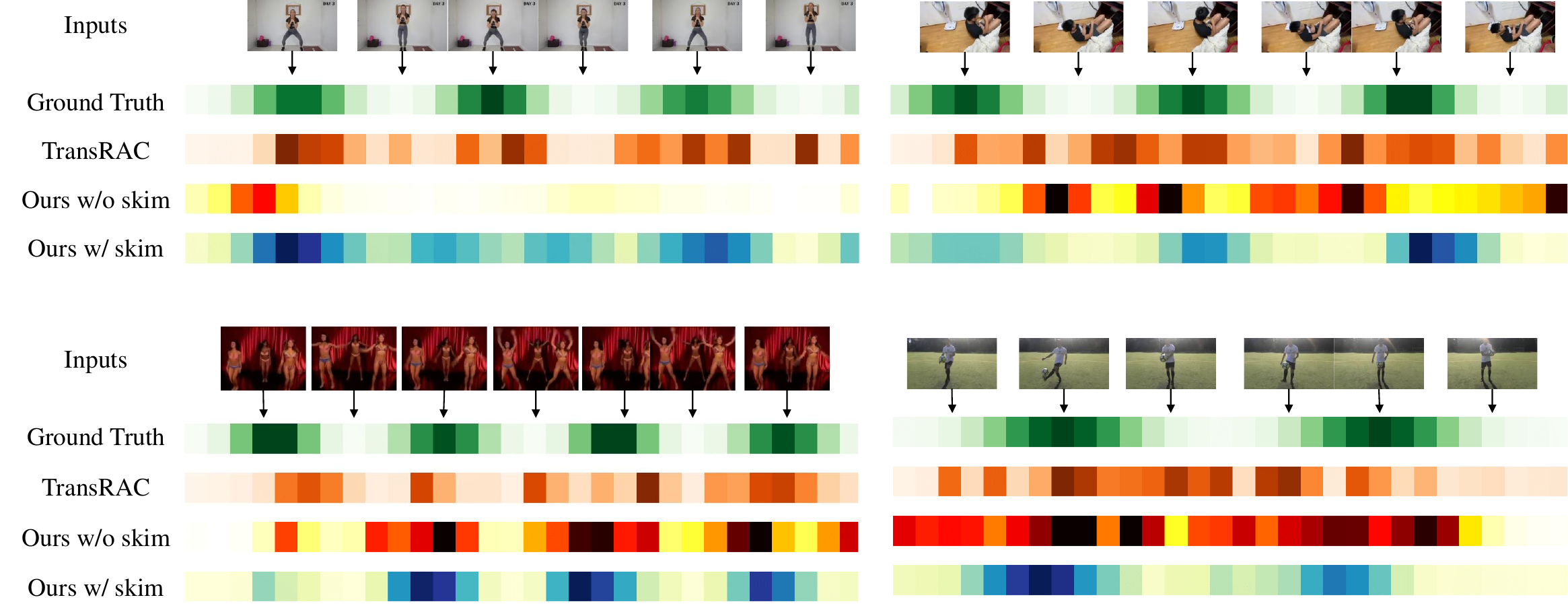}
   \caption{Visualization of ground-truths and predictions. Action classes: ``squat'', ``sit up'', ``jump jacks'', and ``juggle the ball''.  The deeper colors indicate higher action probabilities. Best viewed in color.}
   \label{fig: wave}
   \vspace{5pt}
\end{figure*}

\begin{figure*}[t]
    \centering
    \includegraphics[width=\textwidth]{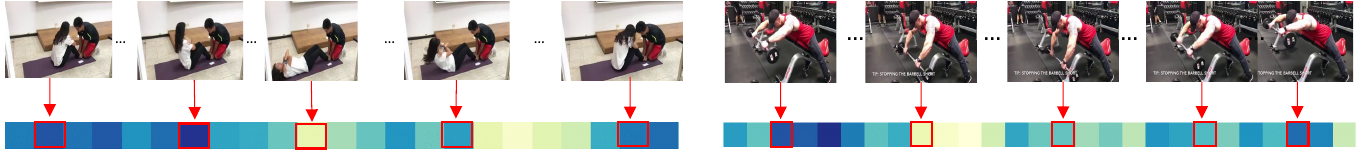}
    \caption{Visualization of attention weights in LSAG. The deeper color indicates higher attention. Action classes: ``sit up'' and ``front raise''. Best viewed in color.}
    \label{fig:visual}
\end{figure*}

\subsection{Qualitative Analysis}
\noindent \textbf{Visualization of predicted density map.} As shown in Fig.\ref{fig: wave}, we visualize the density maps of the ground truths and the predictions. From the visualization, we find that: (1) Compared with TransRAC \cite{hu2022transrac}, our method can locate the action periods more accurately, which differentiates the frames that belong to the target action and the frames that are distractions, thus avoiding unnecessary attention on irrelevant frames. (2) By using the guidance from \textit{skim branch}, it can help the model suppress the wrong focus on the transitional actions, and highlight the focus on the target actions.

\noindent \textbf{Visualization of LSAG.} In Fig.\ref{fig:visual}, the attention weights of LSAG are visualized. We find that the network pays more attention to frames with more useful information about the target action.

\section{Conclusion}
In this paper, we propose a framework for action counting, dubbed SkimFocusNet, which adheres to human intuition to construct a dual-branch architecture. The \textit{skim branch} aims to capture the possible target action and offers guidance to the \textit{focus branch} for action counting. In this way, the model can precisely detect repetitive actions and ignore distracting ones. 

We establish the Multi-RepCount dataset and the problem setting of \textit{specified action counting} to compare the performance of different methods under more complicated conditions.

Our SkimFocusNet is evaluated on RepCount, UCFRep, and Multi-RepCount datasets and achieves state-of-the-art performance over previous methods. Extensive diagnostic experiments and visualizations are given to evaluate the effectiveness of the proposed network.

\section*{Data Availability Statement}
Some of the datasets used in this paper are available online.
RepCount Part-A~\footnote{\url{https://svip-lab.github.io/dataset/RepCount_dataset.html}} and UCFRep~\footnote{\url{https://github.com/Xiaodomgdomg/Deep-Temporal-Repetition-Counting}} can be downloaded from their official website accordingly.
The proposed Multi-RepCount and source code will be available upon acceptance.

{
\bibliographystyle{unsrt}
\bibliography{reference}}

\begin{thebibliography}{10}

\bibitem{kong2022human}
Yu~Kong and Yun Fu.
\newblock Human action recognition and prediction: A survey.
\newblock {\em International Journal of Computer Vision}, 130(5):1366--1401, 2022.

\bibitem{zhang2020context}
Huaidong Zhang, Xuemiao Xu, Guoqiang Han, and Shengfeng He.
\newblock Context-aware and scale-insensitive temporal repetition counting.
\newblock In {\em Proceedings of the IEEE/CVF Conference on Computer Vision and Pattern Recognition}, pages 670--678, 2020.

\bibitem{dwibedi2020counting}
Debidatta Dwibedi, Yusuf Aytar, Jonathan Tompson, Pierre Sermanet, and Andrew Zisserman.
\newblock Counting out time: Class agnostic video repetition counting in the wild.
\newblock In {\em Proceedings of the IEEE/CVF Conference on Computer Vision and Pattern Recognition}, pages 10387--10396, 2020.

\bibitem{hu2022transrac}
Huazhang Hu, Sixun Dong, Yiqun Zhao, Dongze Lian, Zhengxin Li, and Shenghua Gao.
\newblock Transrac: Encoding multi-scale temporal correlation with transformers for repetitive action counting.
\newblock In {\em Proceedings of the IEEE/CVF Conference on Computer Vision and Pattern Recognition}, pages 19013--19022, 2022.

\bibitem{junejo2010view}
Imran~N Junejo, Emilie Dexter, Ivan Laptev, and Patrick Perez.
\newblock View-independent action recognition from temporal self-similarities.
\newblock {\em IEEE Transactions on Pattern Analysis and Machine Intelligence}, 33(1):172--185, 2010.

\bibitem{vaswani2017attention}
Ashish Vaswani, Noam Shazeer, Niki Parmar, Jakob Uszkoreit, Llion Jones, Aidan~N Gomez, {\L}ukasz Kaiser, and Illia Polosukhin.
\newblock Attention is all you need.
\newblock {\em Advances in Neural Information Processing Systems}, 30, 2017.

\bibitem{vlachos2005periodicity}
Michail Vlachos, Philip Yu, and Vittorio Castelli.
\newblock On periodicity detection and structural periodic similarity.
\newblock In {\em Proceedings of the SIAM International Conference on Data Mining}, pages 449--460. SIAM, 2005.

\bibitem{liu2019context}
Weizhe Liu, Mathieu Salzmann, and Pascal Fua.
\newblock Context-aware crowd counting.
\newblock In {\em Proceedings of the IEEE/CVF Conference on Computer Vision and Pattern Recognition}, pages 5099--5108, 2019.

\bibitem{ranjan2018iterative}
Viresh Ranjan, Hieu Le, and Minh Hoai.
\newblock Iterative crowd counting.
\newblock In {\em Proceedings of the European Conference on Computer Vision}, pages 270--285, 2018.

\bibitem{tan2019crowd}
Xin Tan, Chun Tao, Tongwei Ren, Jinhui Tang, and Gangshan Wu.
\newblock Crowd counting via multi-layer regression.
\newblock In {\em Proceedings of the ACM International Conference on Multimedia}, pages 1907--1915, 2019.

\bibitem{wan2019adaptive}
Jia Wan and Antoni Chan.
\newblock Adaptive density map generation for crowd counting.
\newblock In {\em Proceedings of the IEEE/CVF International Conference on Computer Vision}, pages 1130--1139, 2019.

\bibitem{zhang2016single}
Yingying Zhang, Desen Zhou, Siqin Chen, Shenghua Gao, and Yi~Ma.
\newblock Single-image crowd counting via multi-column convolutional neural network.
\newblock In {\em Proceedings of the IEEE Conference on Computer Vision and Pattern Recognition}, pages 589--597, 2016.

\bibitem{zhang20223d}
Qi~Zhang and Antoni~B Chan.
\newblock 3d crowd counting via geometric attention-guided multi-view fusion.
\newblock {\em International Journal of Computer Vision}, 130(12):3123--3139, 2022.

\bibitem{zhang2022wide}
Qi~Zhang and Antoni~B Chan.
\newblock Wide-area crowd counting: Multi-view fusion networks for counting in large scenes.
\newblock {\em International Journal of Computer Vision}, 130(8):1938--1960, 2022.

\bibitem{xu2022autoscale}
Chenfeng Xu, Dingkang Liang, Yongchao Xu, Song Bai, Wei Zhan, Xiang Bai, and Masayoshi Tomizuka.
\newblock Autoscale: Learning to scale for crowd counting.
\newblock {\em International Journal of Computer Vision}, 130(2):405--434, 2022.

\bibitem{albu2008generic}
A~Branzan Albu, Robert Bergevin, and S{\'e}bastien Quirion.
\newblock Generic temporal segmentation of cyclic human motion.
\newblock {\em Pattern Recognition}, 41(1):6--21, 2008.

\bibitem{azy2008segmentation}
Ousman Azy and Narendra Ahuja.
\newblock Segmentation of periodically moving objects.
\newblock In {\em International Conference on Pattern Recognition}, pages 1--4. IEEE, 2008.

\bibitem{cutler2000robust}
Ross Cutler and Larry~S. Davis.
\newblock Robust real-time periodic motion detection, analysis, and applications.
\newblock {\em IEEE Transactions on Pattern Analysis and Machine Intelligence}, 22(8):781--796, 2000.

\bibitem{laptev2005periodic}
Ivan Laptev, Serge~J Belongie, Patrick P{\'e}rez, and Josh Wills.
\newblock Periodic motion detection and segmentation via approximate sequence alignment.
\newblock In {\em IEEE International Conference on Computer Vision}, volume~1, pages 816--823. IEEE, 2005.

\bibitem{lu2004repetitive}
ChunMei Lu and Nicola~J Ferrier.
\newblock Repetitive motion analysis: Segmentation and event classification.
\newblock {\em IEEE Transactions on Pattern Analysis and Machine Intelligence}, 26(2):258--263, 2004.

\bibitem{panagiotakis2018unsupervised}
Costas Panagiotakis, Giorgos Karvounas, and Antonis Argyros.
\newblock Unsupervised detection of periodic segments in videos.
\newblock In {\em IEEE International Conference on Image Processing}, pages 923--927. IEEE, 2018.

\bibitem{pogalin2008visual}
Erik Pogalin, Arnold~WM Smeulders, and Andrew~HC Thean.
\newblock Visual quasi-periodicity.
\newblock In {\em IEEE Conference on Computer Vision and Pattern Recognition}, pages 1--8. IEEE, 2008.

\bibitem{tralie2018quasi}
Christopher~J Tralie and Jose~A Perea.
\newblock (quasi) periodicity quantification in video data, using topology.
\newblock {\em SIAM Journal on Imaging Sciences}, 11(2):1049--1077, 2018.

\bibitem{tsai1994cyclic}
Ping-Sing Tsai, Mubarak Shah, Katharine Keiter, and Takis Kasparis.
\newblock Cyclic motion detection for motion based recognition.
\newblock {\em Pattern recognition}, 27(12):1591--1603, 1994.

\bibitem{briassouli2007extraction}
Alexia Briassouli and Narendra Ahuja.
\newblock Extraction and analysis of multiple periodic motions in video sequences.
\newblock {\em IEEE Transactions on Pattern Analysis and Machine Intelligence}, 29(7):1244--1261, 2007.

\bibitem{burghouts2006quasi}
Gertjan~J Burghouts and J-M Geusebroek.
\newblock Quasi-periodic spatiotemporal filtering.
\newblock {\em IEEE Transactions on Image Processing}, 15(6):1572--1582, 2006.

\bibitem{chetverikov2006motion}
Dmitry Chetverikov and S{\'a}ndor Fazekas.
\newblock On motion periodicity of dynamic textures.
\newblock In {\em Proceedings of the British Machine Vision Conference}, volume~1, pages 167--176. Citeseer, 2006.

\bibitem{davis2000categorical}
James Davis, Aaron Bobick, and Whitman Richards.
\newblock Categorical representation and recognition of oscillatory motion patterns.
\newblock In {\em Proceedings of IEEE Conference on Computer Vision and Pattern Recognition}, volume~1, pages 628--635. IEEE, 2000.

\bibitem{thangali2005periodic}
Ashwin Thangali and Stan Sclaroff.
\newblock Periodic motion detection and estimation via space-time sampling.
\newblock In {\em IEEE Workshops on Applications of Computer Vision}, volume~2, pages 176--182. IEEE, 2005.

\bibitem{runia2019repetition}
Tom~FH Runia, Cees~GM Snoek, and Arnold~WM Smeulders.
\newblock Repetition estimation.
\newblock {\em International Journal of Computer Vision}, 127(9):1361--1383, 2019.

\bibitem{levy2015live}
Ofir Levy and Lior Wolf.
\newblock Live repetition counting.
\newblock In {\em Proceedings of the IEEE International Conference on Computer Vision}, pages 3020--3028, 2015.

\bibitem{runia2018real}
Tom~FH Runia, Cees~GM Snoek, and Arnold~WM Smeulders.
\newblock Real-world repetition estimation by div, grad and curl.
\newblock In {\em Proceedings of the IEEE Conference on Computer Vision and Pattern Recognition}, pages 9009--9017, 2018.

\bibitem{zhang2021repetitive}
Yunhua Zhang, Ling Shao, and Cees~GM Snoek.
\newblock Repetitive activity counting by sight and sound.
\newblock In {\em Proceedings of the IEEE/CVF Conference on Computer Vision and Pattern Recognition}, pages 14070--14079, 2021.

\bibitem{Li_2024_WACV}
Xinjie Li and Huijuan Xu.
\newblock Repetitive action counting with motion feature learning.
\newblock In {\em Proceedings of the IEEE/CVF Winter Conference on Applications of Computer Vision (WACV)}, pages 6499--6508, January 2024.

\bibitem{simonyan2014two}
Karen Simonyan and Andrew Zisserman.
\newblock Two-stream convolutional networks for action recognition in videos.
\newblock {\em Advances in Neural Information Processing Systems}, 27, 2014.

\bibitem{feichtenhofer2019slowfast}
Christoph Feichtenhofer, Haoqi Fan, Jitendra Malik, and Kaiming He.
\newblock Slowfast networks for video recognition.
\newblock In {\em Proceedings of the IEEE/CVF International Conference on Computer Vision}, pages 6202--6211, 2019.

\bibitem{soomro2012ucf101}
Khurram Soomro, Amir~Roshan Zamir, and Mubarak Shah.
\newblock Ucf101: A dataset of 101 human actions classes from videos in the wild.
\newblock {\em arXiv preprint arXiv:1212.0402}, 2012.

\bibitem{feichtenhofer2020x3d}
Christoph Feichtenhofer.
\newblock X3d: Expanding architectures for efficient video recognition.
\newblock In {\em Proceedings of the IEEE/CVF Conference on Computer Vision and Pattern Recognition}, pages 203--213, 2020.

\bibitem{liu2021tam}
Zhaoyang Liu, Limin Wang, Wayne Wu, Chen Qian, and Tong Lu.
\newblock Tam: Temporal adaptive module for video recognition.
\newblock In {\em Proceedings of the IEEE/CVF International Conference on Computer Vision}, pages 13708--13718, 2021.

\bibitem{liu2022video}
Ze~Liu, Jia Ning, Yue Cao, Yixuan Wei, Zheng Zhang, Stephen Lin, and Han Hu.
\newblock Video swin transformer.
\newblock In {\em Proceedings of the IEEE/CVF Conference on Computer Vision and Pattern Recognition}, pages 3202--3211, 2022.

\bibitem{huang2020improving}
Yifei Huang, Yusuke Sugano, and Yoichi Sato.
\newblock Improving action segmentation via graph-based temporal reasoning.
\newblock In {\em Proceedings of the IEEE/CVF Conference on Computer Vision and Pattern Recognition}, pages 14024--14034, 2020.

\bibitem{yi2021asformer}
Fangqiu Yi, Hongyu Wen, and Tingting Jiang.
\newblock Asformer: Transformer for action segmentation.
\newblock {\em arXiv preprint arXiv:2110.08568}, 2021.

\end{thebibliography}

\end{document}